\documentclass{article}

\usepackage{microtype}
\usepackage{graphicx}
\usepackage{subcaption}
\usepackage{booktabs}
\usepackage{tabularx}
\newcolumntype{L}{>{\raggedright\arraybackslash}X}
\newcolumntype{C}{>{\centering\arraybackslash}X}
\usepackage{hyperref}

\usepackage[accepted]{icml2026}

\usepackage{amsmath}
\usepackage{amssymb}
\usepackage{amsfonts}
\usepackage{nicefrac}
\usepackage{float}
\usepackage{multirow}
\usepackage{makecell}
\usepackage{bm}
\usepackage{tcolorbox}
\usepackage{placeins}
\usepackage{stfloats}
\usepackage{caption}
\tcbuselibrary{breakable, skins}

\usepackage[capitalize,noabbrev]{cleveref}

\definecolor{darkblue}{rgb}{0, 0, 0.5}
\hypersetup{colorlinks=true, citecolor=darkblue, linkcolor=darkblue, urlcolor=darkblue}

\icmltitlerunning{Impact of Expert Training Duration on Model Merging}

\begin{document}

\twocolumn[
\icmltitle{Are we Merging the Right Models? Impact of Expert Training Duration on Model Merging for LLMs}

\icmlsetsymbol{equal}{*}

\begin{icmlauthorlist}
\icmlauthor{Nikita Kozodoi}{aws}
\icmlauthor{Zainab Afolabi}{aws}
\icmlauthor{Jack Butler}{aws}
\end{icmlauthorlist}

\icmlaffiliation{aws}{Amazon Web Services}
\icmlcorrespondingauthor{Nikita Kozodoi}{kozodoi@amazon.com}

\vskip 0.3in
]

\printAffiliationsAndNotice{}

\begin{abstract}
Multi-task model merging combines separately trained expert models into a single model that handles all tasks without co-training. Standard practice merges experts at their optimal validation loss. We challenge this convention by systematically studying how training duration of domain experts affects the quality of the merged model. We fine-tune experts on five domains (Math, Code, Instruction Following, Multilingual, and Safety) across three model sizes (Qwen 3.5 0.8B, 2B, and 4B), saving checkpoints from 25\% to 500\% of the optimal training steps and evaluating five merging methods at each duration. Our findings reveal a striking method-dependent pattern: simple averaging degrades sharply with overfitting, while sparsification-based methods achieve their best performance well past the validation optimum. We formalize this through bias-variance decomposition analysis, drawing a parallel to random forests where averaging benefits from high-variance individual learners. These results suggest that training duration and merging method should be chosen jointly rather than independently.
\end{abstract}

%
%

\section{Introduction}
\label{sec:intro}

Model merging has emerged as a practical approach for combining expert models trained separately into a single multi-task model \citep{wortsman2022model, ilharco2023editing}. By operating directly in weight space, merging avoids the computational overhead of multi-task joint training and the data-sharing requirements that come with it. Unlike model ensembling or mixture-of-experts approaches, weight-space merging does not increase inference latency or memory footprint, since all individual experts are merged into a single model. These benefits make model merging particularly attractive for Large Language Models (LLMs), where training costs are substantial, and inference latency must remain low.

A common assumption in the model merging literature is that domain experts should be trained to optimal validation loss before merging. This convention is implicit in major benchmarks and method comparisons: experts are fine-tuned with standard early-stopping or fixed-epoch protocols and then frozen at their best validation checkpoint before being combined \citep{he2025mergebench, yadav2023ties, yu2024language}. The implicit assumption is that each expert contributes its best possible knowledge to the merged model when frozen at its individual optimum. Recent work has begun to question this assumption: \citet{horoi2025less} show, for vision (CLIP) and encoder-decoder (T5) models, that experts trained well past their individual optimum can in fact \emph{hurt} the merged model. We revisit the question for decoder-only LLMs fine-tuned with parameter-efficient adapters and find a more nuanced picture in which the right training duration depends strongly on the merging method.

This paper challenges this assumption. Drawing an analogy from ensemble learning theory, where random forests deliberately grow each decision tree to high depth before averaging predictions \citep{breiman2001random}, we hypothesize that the optimal training duration for merging may differ from the optimal duration for individual model performance. We investigate whether a similar mechanism operates in weight space when merging fine-tuned LLM experts: do some merging methods behave like the averaging step in a random forest and benefit from intentionally overtrained experts?

We focus on small open-weight LLMs because of their prevalence in industrial fine-tuning pipelines. Enterprises commonly fine-tune models below 8B to match or exceed the task-specific quality of larger general-purpose models at a fraction of cost and latency \citep{belcak2025small, abdin2024phi}. Model merging is especially relevant in this regime, since deploying multiple specialized models is typically more expensive than one merged model with comparable capabilities. We conduct a systematic study across 3 model sizes (Qwen 3.5 0.8B, 2B, 4B), 5 task domains (Math, Code, Instruction Following, Multilingual, Safety) and 5 merging methods (Simple Averaging, Task Arithmetic, TIES-Merging, DARE+TIES, Greedy Soup).

Our contributions are as follows. First, we present a systematic study of how training duration of domain experts affects the quality of merged decoder-only LLMs, spanning different model sizes, domains, and merging methods. Second, we show that the optimal training duration is method-dependent: Simple Averaging peaks with undertrained experts, Task Arithmetic and Greedy Soup occupy an intermediate regime, while sparsification-based methods (TIES, DARE+TIES) peak with overfitted experts. We explain these results through bias-variance-covariance decomposition and mode connectivity analysis, showing that overfitted experts provide higher diversity that benefits methods with interference-resolution mechanisms, analogous to how variance reduction enables overtrained trees in random forests. Together, these findings translate into practical guidance: with sparsification-based merging, train each expert past the validation optimum rather than early-stopping.

\section{Related Work}
\label{sec:related}

Weight-space model merging combines separately fine-tuned models without retraining. Model Soups \citep{wortsman2022model} demonstrated that averaging weights of models fine-tuned with different hyperparameters improves accuracy and robustness. Recent work has proposed merging methods that go beyond simple averaging. Task Arithmetic \citep{ilharco2023editing} formalizes task vectors (weight deltas from pre-training) that can be added to compose multi-task models. TIES-Merging \citep{yadav2023ties} addresses interference between task vectors through trimming, sign election, and disjoint merging. DARE \citep{yu2024language} randomly drops and rescales delta parameters before merging. The recent MergeBench suite \citep{he2025mergebench} provides a comprehensive evaluation of 8 merging methods across 5 domains, which we adopt for our experiments. Most of these works fix the merging recipe and produce experts through a standard fine-tuning protocol with early stopping or a fixed budget chosen for individual model quality, treating the choice of training duration as orthogonal to the merging step. ATM \citep{zhou2024atm} instead interleaves tuning and merging, reinterpreting task arithmetic as a single noisy gradient step toward a joint objective, which similarly couples the optimization trajectory to the merge. We instead hold the merging methods fixed and vary expert training duration explicitly, and show that this dimension is in fact tightly coupled to the choice of merging method.

A separate line of work studies weight averaging along a single training trajectory. Stochastic Weight Averaging (SWA) \citep{izmailov2018averaging} averages checkpoints from late training to reach better-generalizing minima, and LAWA \citep{kaddour2022stop} extends this to maintain a FIFO queue of recent checkpoints, which yields gains even when individual checkpoints are past validation optimum. Most directly related to our motivation, Post-Hoc Reversal \citep{ranjan2024posthoc} demonstrates validation-optimal single model checkpoints are not the validation-optimal checkpoints for an ensemble of those models, suggesting that selection criteria designed for individual models can be misleading once aggregation is involved. All of these works study \emph{single-trajectory temporal} averaging within one task. We instead study \emph{cross-task} merging where experts are trained on disjoint data, and find the stopping rule for the merged model.

Most directly related to our work, \citet{horoi2025less} also vary expert training duration across merging methods, and find that overtrained experts degrade merging for CLIP and T5 models under full fine-tuning and LoRA, tracing the effect to the memorization of difficult examples that merging discards. Our study examines the same question for substantially larger decoder-only LLMs fine-tuned with quantized low-rank adapters, and finds that the picture is method-dependent in this regime; we discuss the relationship to their findings in Section~\ref{sec:results}.

DiWA \citep{rame2022diverse} provides a bias-variance-covariance decomposition for weight-averaged models, showing that averaging succeeds when variance dominates the error budget. \citet{tran2026leveraging} explicitly analyze model soups through this decomposition. We apply this framework to understand why overfitted experts may benefit sparsification-based model merging methods.


\section{Experimental Setup}
\label{sec:methods}

We use the Qwen 3.5 model family \citep{qwen35} at three scales: 0.8B, 2B, and 4B total parameters. All experts are fine-tuned with QLoRA \citep{dettmers2024qlora}, i.e.\ low-rank adapters at rank $r=16$ on top of a 4-bit quantized base model, targeting all attention and MLP projection modules. We use a constant learning rate of $2 \times 10^{-4}$ with a 100-step linear warmup, an effective batch size of 8, and the AdamW optimizer. The constant schedule with no decay ensures the learning rate remains active past the optimal validation loss checkpoint, allowing each expert to overfit.

Following the setup of MergeBench \citep{he2025mergebench}, we adopt 5 task domains, each with a dedicated fine-tuning dataset and evaluation benchmark. For each dataset, we fine-tune a Qwen 3.5 expert using only the training data from that domain. All individual experts are then merged using all 5 domain LoRA adapters. We aggregate metrics over the 4 domains with accuracy metrics; the Safety domain measures the refuse-to-answer rate (higher is better) and is reported in Section~\ref{sec:safety}.
For each domain and model size, we identify the validation optimum T* independently, defined as the training step with minimum loss on that domain's held-out validation split (1K examples). Because T* is selected separately per domain and size, the absolute number of steps it corresponds to varies across experts, so all training durations are expressed as multiples of each expert's own T* rather than as absolute step counts. We continue training up to 5$\times$ T* and save LoRA adapters at 8 checkpoints: \{0.25, 0.5, 0.75, 1.0, 1.5, 2.0, 3.0, 5.0\}$\times$ T*, spanning from heavily undertrained to extremely overfitted experts. 

\begin{table*}[!b]
\centering
\small
\begin{tabularx}{\textwidth}{lLLL}
\toprule
Domain & Training Data & Evaluation Benchmark & Metric \\
\midrule
Mathematics & DART-Math \citep{wei2024dartmath} & GSM8K \citep{cobbe2021gsm8k} & Exact-match accuracy \\
Code & Magicoder \citep{luo2024magicoder} & HumanEval \citep{chen2021humaneval} & Pass@1 \\
Instruction & TULU-3 \citep{lambert2024tulu3} & IFEval \citep{zhou2023ifeval} & Prompt-level accuracy \\
Multilingual & Aya \citep{singh2024aya} & ARC \citep{clark2018arc} & Normalized accuracy \\
Safety & BeaverTails \citep{ji2024beavertails} & HarmBench \citep{mazeika2024harmbench} & Refuse-to-answer rate \\
\bottomrule
\end{tabularx}
\vspace{0.5ex}
\caption{Task domains, datasets, and benchmarks. Training datasets are subsampled to 10K training and 1K validation examples.}
\label{tab:domains}
\end{table*}

We evaluate 5 merging methods spanning different families: \textbf{Simple Averaging} \citep{wortsman2022model}: equal-weight linear combination of experts ($w_i = 1/N$), \textbf{Task Arithmetic} \citep{ilharco2023editing}: scales task vectors by $\lambda$ before combining with $\lambda \in \{0.5, 1.0, 1.5\}$ selected via grid search, \textbf{TIES} \citep{yadav2023ties}: trims small-magnitude parameters, resolves sign conflicts, merging values with density $k=0.5$, \textbf{DARE+TIES} \citep{yu2024language}: randomly drops and rescales delta parameters before TIES merging with density $k=0.5$ and \textbf{Greedy Soup} \citep{wortsman2022model}: iteratively adds experts to the merge, keeping each only if it improves the held-out score of the merged model.

\section{Results}
\label{sec:results}

Here we first show results for the average multi-domain accuracy of merged models as a function of training duration, broken down by merging method and model size. We aggregate the quality scores over the four domains: Math, Code, Instruction, Multilingual. Full results are provided in Appendix~\ref{app:full}. Then, we proceed to show results on refuse-to-answer(\%) on HarmBench across training durations.

\subsection{Training duration vs. merged model quality}

\begin{figure*}[!t]
\centering
\includegraphics[width=0.95\textwidth, trim=0 0 0 1.5cm, clip]{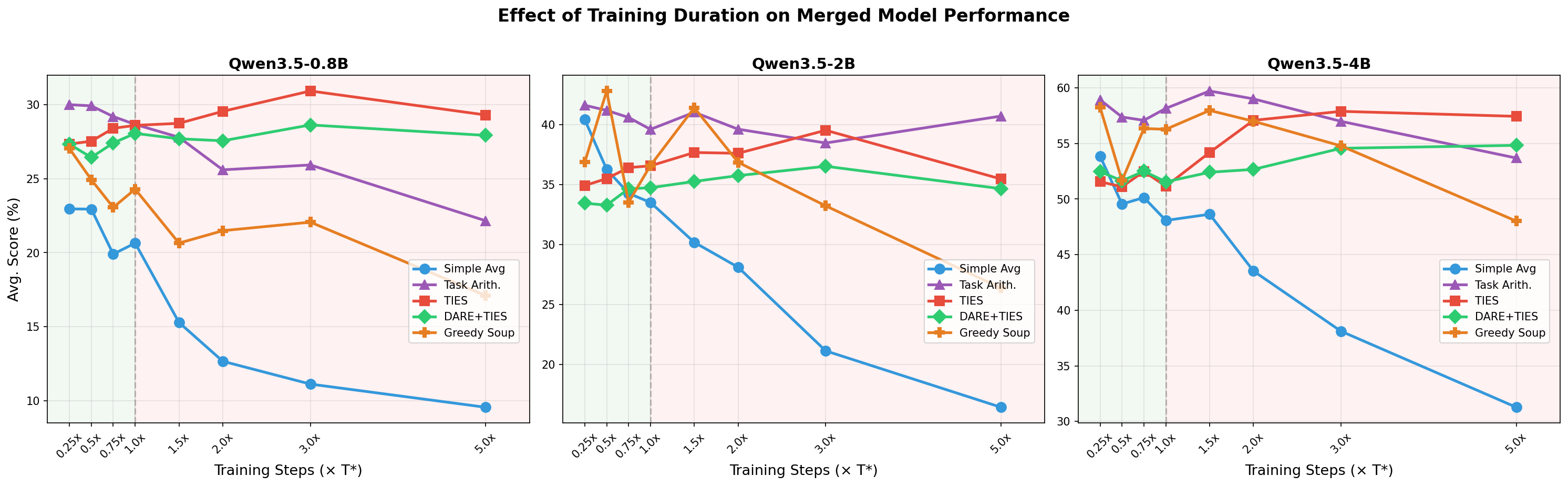}
\caption{Effect of training duration on merged model performance. Each line represents one merging method; the x-axis shows training steps as a multiple of T* (optimal validation checkpoint per expert). Green shading indicates the undertrained regime ($<$T*); red indicates the overfitted regime ($>$T*). Results are averaged over the four accuracy-based domains. Optimal T steps is shown as method-specific.
}
\label{fig:hero}
\end{figure*}

The results in Figure~\ref{fig:hero} reveal a striking method-dependent pattern. Simple Averaging achieves its best performance in the undertrained regime (0.25--0.75$\times$ T*) and degrades with longer training, losing 14--23 percentage points on average by 5$\times$ T* (and up to 45 points on individual domains such as Math). In contrast, TIES-Merging peaks well past T* (at 1.5--3$\times$ T* for 0.8B and 2B, and 3--5$\times$ T* for 4B), exceeding its T* score by 2.5--8.7 percentage points. DARE+TIES shows a similar but flatter profile, remaining stable across training durations with a slight preference for overfitted experts. Task Arithmetic occupies an intermediate regime: it degrades on some domains (especially Math, where it peaks at 0.25$\times$ T*) but tolerates moderate overfitting on others. Greedy Soup, which selectively includes experts based on validation improvement, tends to favor undertrained experts for smaller LLM sizes but shows a more balanced selection at 4B. Its curve is notably less smooth than the other methods because the greedy selection includes different subsets of experts at each training duration.

We attribute this discrepancy to how each method handles inter-expert interference. Simple Averaging treats every parameter of every expert with equal weight, so any noisy or specialized parameter in an overtrained expert directly pollutes the merged adapter. Task Arithmetic applies a global scale $\lambda$ to the task vectors before combining, which partially attenuates overfitting-induced noise but cannot resolve sign conflicts or selectively suppress interfering parameters, placing it between Simple Averaging and sparsification methods. TIES and DARE+TIES, by contrast, include explicit interference-resolution steps: TIES trims small-magnitude updates and resolves sign conflicts, while DARE first stochastically prunes deltas and rescales the survivors. These steps act as a variance-reduction mechanism over the experts: when overfitting introduces idiosyncratic, low-magnitude or sign-conflicting components, sparsification removes them before the average is taken. The result is the same intuition that motivates random forests \citep{breiman2001random}: deeper, higher-variance individual learners are tolerated when the aggregation step explicitly suppresses the noise they introduce. Greedy Soup occupies a middle ground because its selection criterion is the held-out validation score of the merge, not of individual experts, so it implicitly co-optimizes training duration and merging composition, but only across a small discrete set of choices.

These results partly agree with and partly diverge from \citet{horoi2025less}, who study the same question on smaller CLIP and T5 models. We agree that averaging-style methods (Simple Averaging, Greedy Soup) favor undertrained experts, but diverge for the sparsification-based methods: where they report that TIES and DARE degrade with overtraining, we find that TIES and DARE+TIES improve. This divergence is consistent with their own observation that a higher LoRA rank attenuates the overtraining penalty, since we use a larger rank ($r=16$ vs.\ $r=8$) on substantially larger decoder-only models with quantized adapters and a constant learning rate.

\subsection{Training duration vs. merged model safety}
\label{sec:safety}

The results in Table~\ref{tab:safety} echo the accuracy pattern with one notable shift. Simple Averaging diffuses the safety signal across all parameters and peaks near T* (1$\times$ for 2B and 4B, 2$\times$ for 0.8B), leaving the merged model at or below 25\% refusal even at 4B. Task Arithmetic preserves more safety than Simple Averaging and shows a clear preference for longer training at larger scales: it peaks at 2$\times$ for 0.8B (10\%) but reaches 30.5\% at 3$\times$ for 2B and 50\% at 3$\times$ for 4B. TIES and DARE+TIES retain the most refusal behavior and, like in the accuracy setting, prefer overtrained experts, with the optimum drifting later as model size grows. DARE+TIES peaks at 1.5$\times$ for 0.8B and 5$\times$ for both 2B and 4B. The 4B TIES merge reaches 68.5\% refusal at 3$\times$, more than 2.5$\times$ higher than Simple Averaging at any duration.

\begin{table}[!b]
\centering
\small
\setlength{\tabcolsep}{3pt}
\resizebox{\columnwidth}{!}{%
\begin{tabular}{lcccccccc}
\toprule
Method & 0.25$\times$ & 0.5$\times$ & 0.75$\times$ & 1.0$\times$ & 1.5$\times$ & 2.0$\times$ & 3.0$\times$ & 5.0$\times$ \\
\midrule
\multicolumn{9}{c}{\textit{Qwen3.5-0.8B}} \\
\midrule
Simple Avg. & 2.5 & 6.5 & 4.5 & 7.0 & 5.0 & \textbf{8.5} & 3.5 & 4.5 \\
Task Arithm. & 5.0 & 5.5 & 4.5 & 9.0 & 5.0 & \textbf{10.0} & 8.0 & 7.0 \\
TIES & 22.5 & 21.5 & 19.0 & 20.0 & \textbf{24.0} & 15.5 & 17.5 & 21.0 \\
DARE+TIES & 21.0 & 22.0 & 20.5 & 19.0 & \textbf{23.5} & 22.5 & 21.0 & 19.5 \\
Greedy Soup & 2.5 & 5.0 & 7.0 & 6.0 & 4.0 & \textbf{9.0} & 5.5 & 2.0 \\
\midrule
\multicolumn{9}{c}{\textit{Qwen3.5-2B}} \\
\midrule
Simple Avg. & 5.0 & 8.5 & 10.0 & \textbf{12.5} & 11.0 & 10.0 & 8.5 & 3.0 \\
Task Arithm. & 9.5 & 13.0 & 18.0 & 30.0 & 26.5 & 29.5 & \textbf{30.5} & 30.0 \\
TIES & 35.0 & 35.0 & 33.5 & \textbf{36.0} & 27.0 & 20.5 & 16.5 & 23.5 \\
DARE+TIES & 34.0 & 34.0 & 37.0 & 37.5 & 41.0 & 36.0 & 40.5 & \textbf{45.5} \\
Greedy Soup & 3.5 & 9.0 & 11.0 & \textbf{12.5} & 10.0 & 9.5 & 9.0 & 2.5 \\
\midrule
\multicolumn{9}{c}{\textit{Qwen3.5-4B}} \\
\midrule
Simple Avg.& 24.0 & 20.5 & 21.5 & \textbf{25.5} & 22.0 & 14.0 & 17.0 & 6.0 \\
Task Arithm. & 49.0 & 44.0 & 48.5 & 42.0 & 46.0 & 39.0 & \textbf{50.0} & 44.0 \\
TIES & 51.5 & 51.0 & 57.5 & 53.5 & 60.5 & 64.0 & \textbf{68.5} & 64.0 \\
DARE+TIES & 50.0 & 49.0 & 53.0 & 53.5 & 55.0 & 55.0 & 60.0 & \textbf{61.5} \\
Greedy Soup & 22.5 & 20.0 & 22.5 & \textbf{23.0} & 21.0 & 22.5 & 17.0 & 7.0 \\
\bottomrule
\end{tabular}%
}
\vspace{0.5ex}
\caption{Refuse-to-answer rate (\%) on HarmBench across training durations and model sizes. Best per row in bold.}
\label{tab:safety}
\end{table}

Greedy Soup behaves similarly to Simple Averaging on safety: peaking near T* (1$\times$ for 2B and 4B at 12.5\% and 23\% respectively, 2$\times$ for 0.8B at 9\%) and retains little refusal at extreme durations. This is because greedy selection optimizes for accuracy-based validation scores, which may actively exclude the safety expert when it hurts accuracy, removing the refusal signal. This is consistent with the interpretation that the safety expert produces a localized parameter update (refusal), and sparsification preserves this signal even when other experts are overtrained, so the same recipe which helps accuracy also strengthens safety.

\section{Analysis}
\label{sec:analysis}

\subsection{Bias-variance decomposition}

To understand why overfitted experts benefit sparsification-based merging, we apply the bias-variance-covariance decomposition from DiWA \citep{rame2022diverse} to our setting. For each training duration, we decompose the merged model's expected error into three components: \textbf{Bias}: how far the average expert prediction is from ground truth which decreases with training, \textbf{Variance}: how much individual expert predictions differ which increases with overfitting and \textbf{Covariance}: the correlation of experts' errors.

\begin{figure*}[t]
\centering
\includegraphics[width=0.98\textwidth, trim=0 0 0 1.5cm, clip]{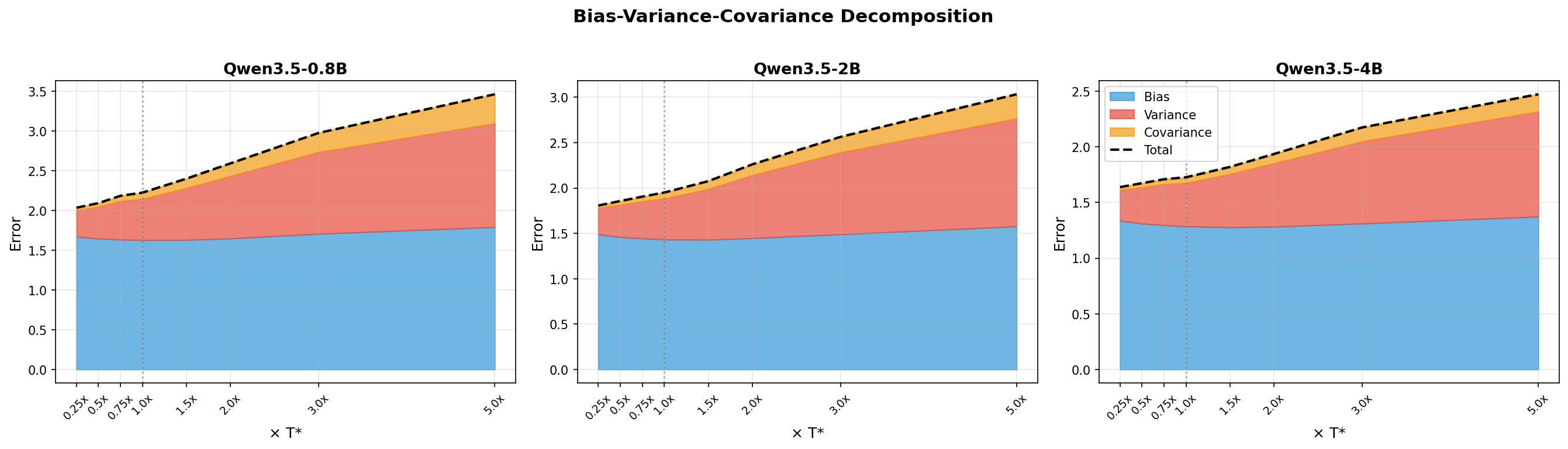}
\caption{Bias-variance-covariance decomposition of the merged model error against training duration, shown across all three model sizes. As training increases past T*, bias drops while variance and covariance rise.}
\label{fig:biasvar}
\end{figure*}

Figure~\ref{fig:biasvar} shows the decomposition across all three model sizes, which clarifies the mechanism behind the method-dependent pattern. As training progresses past T*, bias drops (experts capture task-specific patterns) while variance rises (experts diverge). Simple averaging passes increased variance and covariance in the merged model, leading to degraded performance. Sparsification methods explicitly filter out small or sign-conflicting parameter values, reducing variance and covariance while preserving the low-bias signal. Sparsification acts as a variance-reduction mechanism analogous to the averaging step in random forests, where the aggregation step suppresses idiosyncrasies of deeper, higher-variance trees.

\subsection{Mode connectivity}

We evaluate linear mode connectivity between expert pairs to assess whether overfitted experts remain in the same loss basin. For each pair $(\theta_A, \theta_B)$, we interpolate $\theta(\alpha) = \alpha \theta_A + (1-\alpha)\theta_B$ for $\alpha \in [0,1]$ and measure the loss barrier (maximum loss along the path minus the average of the endpoint losses).

\begin{figure}[t]
\centering
\includegraphics[width=0.98\linewidth]{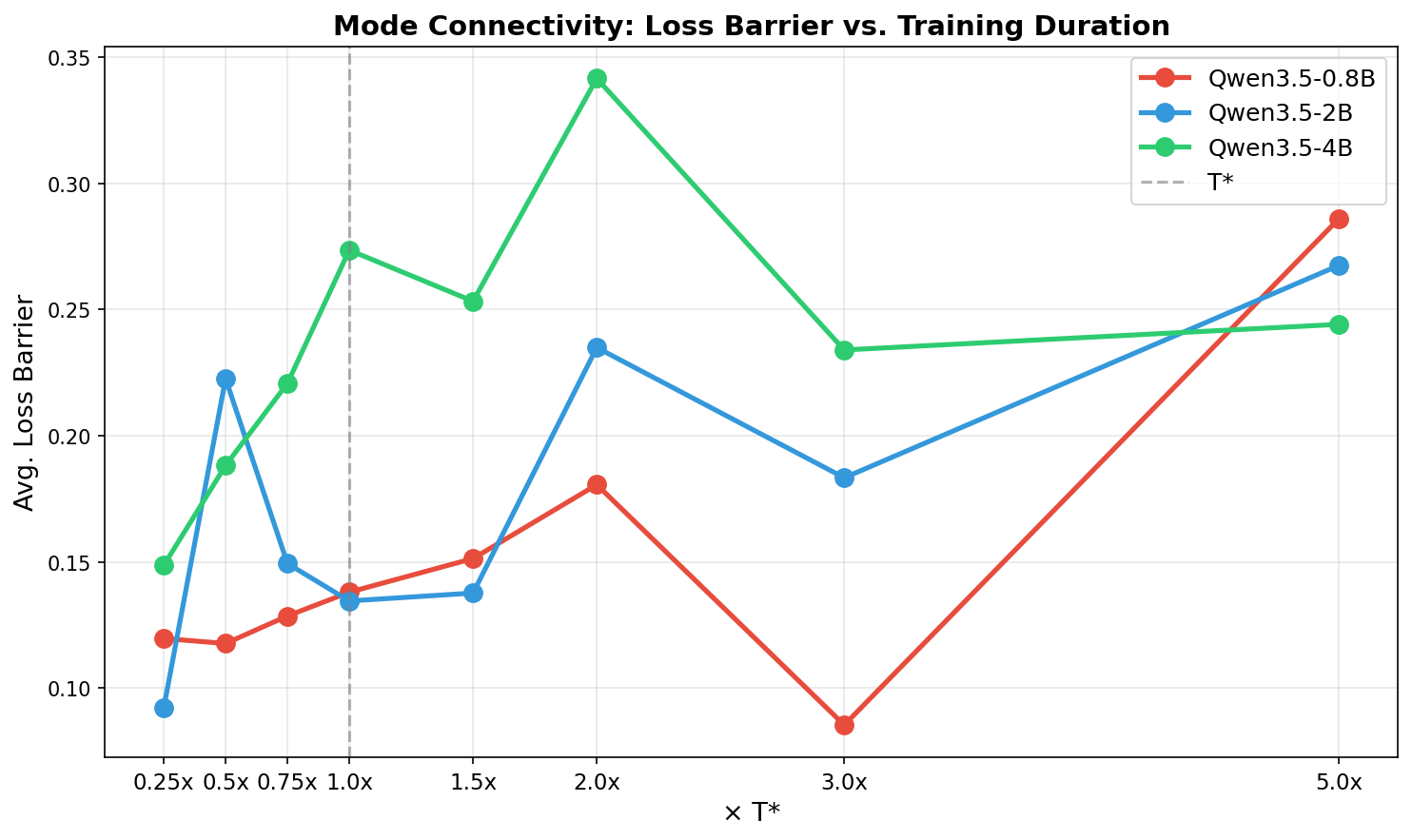}
\caption{Mode connectivity: average loss barrier between expert pairs vs.\ training duration. The barrier grows monotonically with training duration but remains moderate even at 5$\times$ T*.}
\label{fig:modeconn}
\vspace*{-0.42cm}
\end{figure}

Figure~\ref{fig:modeconn} shows that the loss barrier increases with training duration, indicating that overfitted experts are progressively less linearly mode-connected. However, the barrier remains moderate (below 0.35) even at 5$\times$ T*, suggesting that LoRA's low-rank constraint prevents experts from leaving the pre-trained basin. This explains why weight-space merging remains viable even for heavily overfitted experts: the limiting factor is not basin escape but the merging method's ability to suppress expert-specific noise within the basin. In other words, LoRA gives us a wide ``safe zone'' where the mode connectivity breakdown is unlikely.

\section{Conclusion}
\label{sec:conclusion}

We conduct a systematic study of how training duration of domain experts affects the quality of merged multi-task LLMs. Our experiments across three model sizes, five domains and five merging methods reveal that the optimal training duration for merging is fundamentally method-dependent. Simple Averaging benefits from undertrained experts (0.25--1$\times$ T*), Task Arithmetic and Greedy Soup occupy an intermediate regime (0.25--3$\times$ T* depending on the domain), while sparsification-based methods (TIES, DARE+TIES) achieve peak performance with deliberately overfitted experts (2--5$\times$ T*). This finding directly challenges the widespread assumption that domain experts should be merged at their individual validation optima.

Through bias-variance decomposition, we show that overfitted experts provide higher diversity (lower bias, higher variance), which benefits methods equipped with interference-resolution mechanisms. The sparsification step in TIES and DARE acts as a variance-reduction mechanism analogous to the aggregation step in random forests, filtering overfitting-induced noise while preserving task-specific knowledge.

Our results have direct practical implications: when using sparsification-based merging methods, practitioners should deliberately train experts past their validation optimum. For the Qwen 3.5 family, training to 2--5$\times$ T* with TIES or DARE+TIES yields the best merged models across all tested sizes, recovering most of the gap to single-domain experts while retaining all capabilities in a single adapter.
Our study has several limitations. We use a single open-weight model family (Qwen 3.5) and a single fine-tuning recipe: QLoRA adapters at a fixed rank ($r=16$) on a 4-bit quantized base, trained with a constant learning rate. Each of these is potentially load-bearing: prior work finds that full fine-tuning and LoRA respond differently to training budget \citep{horoi2025less}, and learning-rate decay regulates how sharply a model overtrains \citep{rofin2026learning}, so the constant schedule may amplify the regime we study. Our non-deterministic methods (Greedy Soup and DARE's stochastic pruning) are run with a single seed, and the bias-variance-covariance account in Section~\ref{sec:analysis} is an explanatory hypothesis rather than a proven mechanism. Future work should test whether the pattern persists under full fine-tuning, learning-rate decay, larger base models, and other PEFT methods.

\clearpage

%
%

\bibliographystyle{icml2026}
\bibliography{references}

%
%

\clearpage
\appendix

\raggedbottom

\section{Additional Results}
\label{app:full}

\subsection{Main results tables}
\label{app:main_tables}

Tables~\ref{tab:main_08b}--\ref{tab:main_4b} report per-domain accuracy at T* (standard practice) and the best score across all training durations for each merging method and model size. For every method and size, the best training duration outperforms T*.

\begin{table*}[p]
\centering
\small
\setlength{\tabcolsep}{4pt}
\begin{tabular}{lccccc}
\toprule
Model & Mathematics & Code & Instruction & Multilingual & Average \\
\midrule
Base (no FT) & 10.5 & 23.2 & 33.6 & 37.2 & 26.1 \\
Expert @ T* & 38.2 & 23.2 & 29.0 & 39.7 & 32.5 \\
\midrule
Simple Avg @ T* & 8.0 & 17.1 & 18.5 & 39.1 & 20.6 \\
\textbf{Best Simple Averaging} & \textbf{16.8} & \textbf{17.7} & \textbf{18.5} & \textbf{41.3} & \textbf{23.6} \\
Task Arithmetic @ T* & 24.3 & 24.4 & 23.1 & 42.8 & 28.7 \\
\textbf{Best Task Arithmetic} & \textbf{28.6} & \textbf{24.4} & \textbf{27.0} & \textbf{43.6} & \textbf{30.9} \\
TIES @ T* & 18.2 & 22.0 & 33.6 & 40.6 & 28.6 \\
\textbf{Best TIES} & \textbf{22.9} & \textbf{24.4} & \textbf{35.9} & \textbf{41.1} & \textbf{31.1} \\
DARE+TIES @ T* & 16.5 & 21.3 & 35.1 & 39.2 & 28.1 \\
\textbf{Best DARE+TIES} & \textbf{18.0} & \textbf{23.2} & \textbf{35.5} & \textbf{39.8} & \textbf{29.1} \\
Greedy Soup @ T* & 7.9 & 20.1 & 29.6 & 39.6 & 24.3 \\
\textbf{Best Greedy Soup} & \textbf{12.6} & \textbf{24.4} & \textbf{31.2} & \textbf{41.0} & \textbf{27.3} \\
\bottomrule
\end{tabular}
\vspace{0.5ex}
\caption{Qwen3.5-0.8B: per-domain accuracy at T* and best across all training durations (bold).}
\label{tab:main_08b}
\end{table*}

\begin{table*}[p]
\centering
\small
\setlength{\tabcolsep}{4pt}
\begin{tabular}{lccccc}
\toprule
Model & Mathematics & Code & Instruction & Multilingual & Average \\
\midrule
Base (no FT) & 21.8 & 39.0 & 30.1 & 41.7 & 33.2 \\
Expert @ T* & 52.9 & 36.6 & 32.0 & 47.3 & 42.2 \\
\midrule
Simple Averaging @ T* & 23.5 & 32.3 & 27.9 & 50.3 & 33.5 \\
\textbf{Best Simple Averaging} & \textbf{43.6} & \textbf{36.6} & \textbf{30.5} & \textbf{51.5} & \textbf{40.5} \\
Task Arithmetic @ T* & 32.9 & 39.0 & 35.5 & 50.9 & 39.6 \\
\textbf{Best Task Arithmetic} & \textbf{43.1} & \textbf{43.9} & \textbf{35.5} & \textbf{52.3} & \textbf{43.7} \\
TIES @ T* & 28.6 & 40.2 & 32.0 & 45.4 & 36.5 \\
\textbf{Best TIES} & \textbf{30.7} & \textbf{43.3} & \textbf{35.7} & \textbf{48.4} & \textbf{39.5} \\
DARE+TIES @ T* & 27.7 & 37.2 & 30.7 & 43.3 & 34.7 \\
\textbf{Best DARE+TIES} & \textbf{30.0} & \textbf{40.9} & \textbf{31.2} & \textbf{45.9} & \textbf{37.0} \\
Greedy Soup @ T* & 28.9 & 39.0 & 33.5 & 45.0 & 36.6 \\
\textbf{Best Greedy Soup} & \textbf{50.0} & \textbf{39.6} & \textbf{35.9} & \textbf{50.9} & \textbf{44.1} \\
\bottomrule
\end{tabular}
\vspace{0.5ex}
\caption{Qwen3.5-2B: per-domain accuracy at T* and best across all training durations (bold).}
\label{tab:main_2b}
\end{table*}

\begin{table*}[p]
\centering
\small
\setlength{\tabcolsep}{4pt}
\begin{tabular}{lccccc}
\toprule
Model & Mathematics & Code & Instruction & Multilingual & Average \\
\midrule
Base (no FT) & 61.6 & -- & 30.9 & 54.1 & 48.8 \\
Expert @ T* & 78.0 & 59.8 & 44.0 & 54.9 & 59.2 \\
\midrule
Simple Averaging @ T* & 45.6 & 52.4 & 33.8 & 60.4 & 48.1 \\
\textbf{Best Simple Averaging} & \textbf{64.8} & \textbf{56.1} & \textbf{37.0} & \textbf{61.1} & \textbf{54.7} \\
Task Arithmetic @ T* & 66.4 & 59.1 & 46.4 & 60.7 & 58.2 \\
\textbf{Best Task Arithmetic} & \textbf{74.1} & \textbf{65.9} & \textbf{46.4} & \textbf{62.8} & \textbf{62.3} \\
TIES @ T* & 54.9 & 56.7 & 36.6 & 56.4 & 51.1 \\
\textbf{Best TIES} & \textbf{66.7} & \textbf{61.6} & \textbf{51.6} & \textbf{59.4} & \textbf{59.8} \\
DARE+TIES @ T* & 60.3 & 56.1 & 34.2 & 55.6 & 51.5 \\
\textbf{Best DARE+TIES} & \textbf{63.6} & \textbf{61.6} & \textbf{38.1} & \textbf{57.1} & \textbf{55.1} \\
Greedy Soup @ T* & 65.4 & 57.3 & 45.1 & 57.3 & 56.3 \\
\textbf{Best Greedy Soup} & \textbf{66.0} & \textbf{64.6} & \textbf{55.6} & \textbf{58.3} & \textbf{61.1} \\
\bottomrule
\end{tabular}
\vspace{0.5ex}
\caption{Qwen3.5-4B: per-domain accuracy at T* and best across all training durations (bold).}
\label{tab:main_4b}
\end{table*}

\subsection{Per-domain breakdown}
\label{app:perdomain}

\begin{figure*}[b]
\centering
\includegraphics[width=\textwidth]{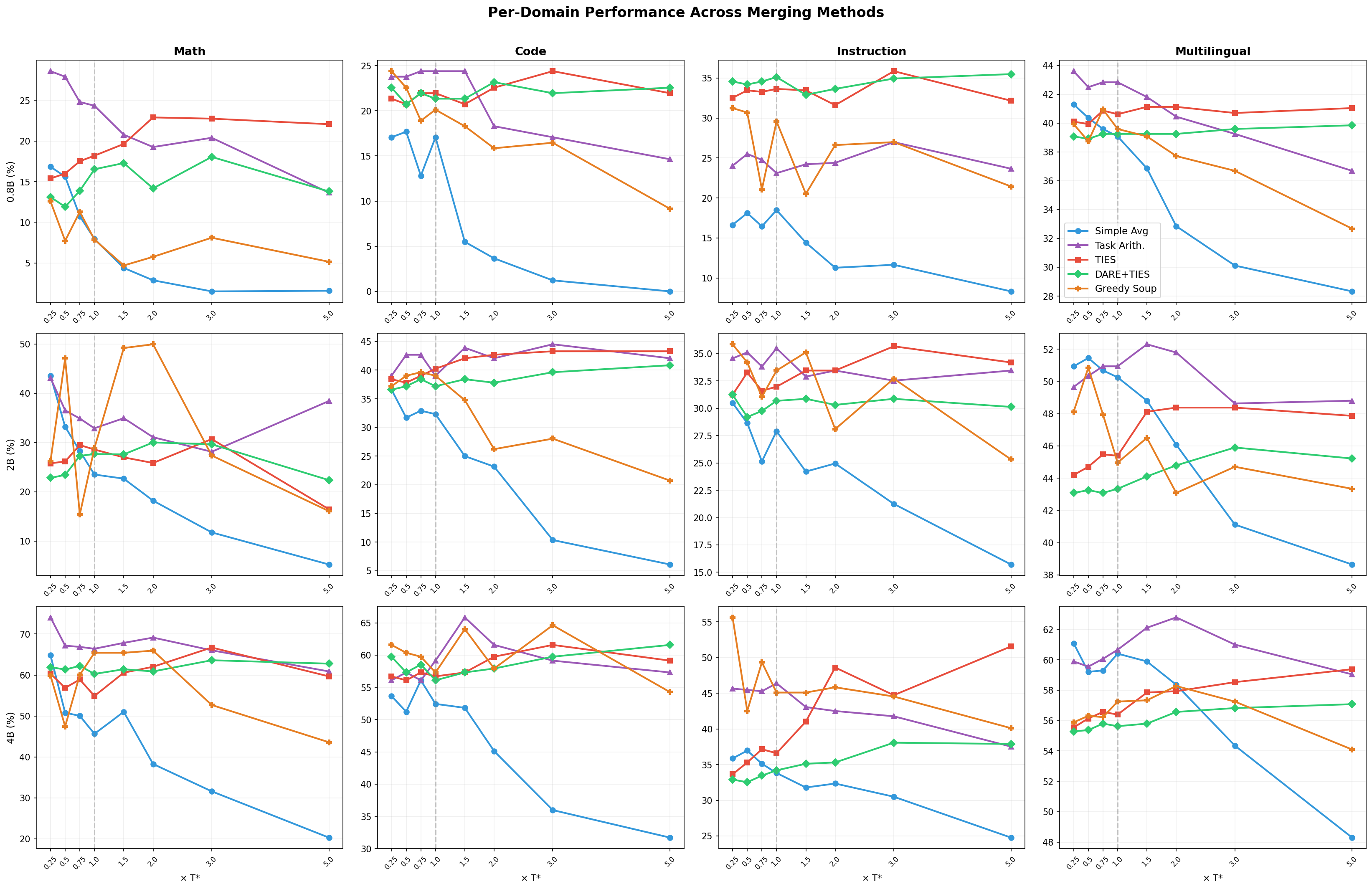}
\caption{Per-domain performance across merging methods, model sizes, and domains. The method-dependent pattern from Figure~\ref{fig:hero} holds across domains, though the magnitude varies.}
\label{fig:grid}
\end{figure*}

Figure~\ref{fig:grid} decomposes the aggregate trends from Figure~\ref{fig:hero} by domain and model size. The method-dependent pattern from the main results holds consistently across domains.

Mathematics shows the sharpest divergence between methods: at the 4B scale, Simple Averaging drops from 64.8\% at 0.25$\times$ T* to 20.3\% at 5$\times$ T*, while TIES improves from 60.3\% at 0.25$\times$ T* to a peak of 66.7\% at 3$\times$ T*. Code generation shows similar trends but with smaller gaps, since HumanEval is more saturated. Instruction Following and Multilingual tasks are more stable across methods, suggesting that the sensitivity to training duration scales with the difficulty of the underlying task: tasks where individual experts gain a lot from extra training are also tasks where merging method choice matters most.

\subsection{Base model scale dependence}
\label{app:scale}

The interaction between training duration and merging quality varies with model size in a way consistent with our bias-variance results. For Simple Averaging, 0.8B models degrade fastest with overfitting (the merged 0.8B model loses essentially all of its math performance by 3$\times$ T*), 2B models are intermediate, and 4B models show the most gradual decline (e.g., from 64.8\% at 0.25$\times$ to 20.3\% at 5$\times$ on Math). For TIES, the optimal overfitting level shifts rightward with model size: 0.8B peaks at 1.5--3$\times$ T*, 2B peaks at 2--3$\times$ T*, and 4B peaks at 3--5$\times$ T*. Table~\ref{tab:scale} summarizes the range of optimal training durations for each method across the four accuracy domains.

\begin{table}[h]
\centering
\small
\setlength{\tabcolsep}{3pt}
\resizebox{\columnwidth}{!}{%
\begin{tabular}{lccc}
\toprule
Method & Qwen 3.5-0.8B & Qwen 3.5-2B & Qwen 3.5-4B \\
\midrule
Simple Averaging & 0.25--1.0$\times$ & 0.25--0.5$\times$ & 0.25--0.75$\times$ \\
Task Arithmetic & 0.25--3.0$\times$ & 0.25--3.0$\times$ & 0.25--2.0$\times$ \\
TIES & 1.5--3.0$\times$ & 2.0--3.0$\times$ & 3.0--5.0$\times$ \\
DARE+TIES & 2.0--5.0$\times$ & 0.25--5.0$\times$ & 3.0--5.0$\times$ \\
Greedy Soup & 0.25--0.75$\times$ & 0.25--2.0$\times$ & 0.25--3.0$\times$ \\
\bottomrule
\end{tabular}%
}
\vspace{0.5ex}
\caption{Range of optimal training durations (multiples of T*) per method across the four accuracy domains, by model size. The optimum for sparsification-based methods shifts rightward as model size grows, while Simple Averaging stays anchored to the undertrained regime regardless of scale.}
\label{tab:scale}
\end{table}

This is consistent with the overparameterization view of fine-tuning. Larger models have more capacity to absorb task-specific knowledge into low-magnitude parameter subsets that are spread across the network, rather than concentrated in a few salient directions. As a result, the additional variance introduced by overtraining is also more diffuse, which makes it easier for sparsification-based methods to filter it out: the noisy directions are still individually small and sign-conflicting after the optimum, just as they need to be for TIES and DARE to discard them. Conversely, in smaller models, additional training pushes a relatively small set of parameters further from the shared basin, and even sparsification cannot fully recover the merged signal once those salient directions disagree across experts. The net effect is that the band of training durations for which sparsification-based merging is helpful both widens and shifts later as model size grows, which provides a useful pattern for practitioners: larger backbones reward more aggressive overtraining of each expert.

\subsection{Optimal training duration}
\label{app:optimal}

Tables~\ref{tab:optimal_08b}--\ref{tab:optimal_4b} report the optimal training duration per domain for each merging method, separately for the 0.8B, 2B, and 4B Qwen3.5 models. They show that the same dichotomy holds across model sizes: averaging-style methods peak undertrained and sparsification-based methods peak overtrained. The Safety column reports the training duration at which each method achieves its highest refuse-to-answer rate.

\begin{table}[h]
\centering
\small
\setlength{\tabcolsep}{3pt}
\resizebox{\columnwidth}{!}{%
\begin{tabular}{lccccc}
\toprule
Method & Mathematics & Code & Instruction & Multilingual & Safety \\
\midrule
Simple Averaging & 0.25$\times$ & 0.5$\times$ & 1.0$\times$ & 0.25$\times$ & 2.0$\times$ \\
Task Arithmetic & 0.25$\times$ & 0.75$\times$ & 3.0$\times$ & 0.25$\times$ & 2.0$\times$ \\
TIES & 2.0$\times$ & 3.0$\times$ & 3.0$\times$ & 1.5$\times$ & 1.5$\times$ \\
DARE+TIES & 3.0$\times$ & 2.0$\times$ & 5.0$\times$ & 5.0$\times$ & 1.5$\times$ \\
Greedy Soup & 0.25$\times$ & 0.25$\times$ & 0.25$\times$ & 0.75$\times$ & 2.0$\times$ \\
\bottomrule
\end{tabular}%
}
\vspace{0.5ex}
\caption{Optimal training duration per domain for Qwen3.5-0.8B.}
\label{tab:optimal_08b}
\end{table}

\begin{table}[h]
\centering
\small
\setlength{\tabcolsep}{3pt}
\resizebox{\columnwidth}{!}{%
\begin{tabular}{lccccc}
\toprule
Method & Mathematics & Code & Instruction & Multilingual & Safety \\
\midrule
Simple Averaging & 0.25$\times$ & 0.25$\times$ & 0.25$\times$ & 0.5$\times$ & 1.0$\times$ \\
Task Arithmetic & 0.25$\times$ & 3.0$\times$ & 1.0$\times$ & 1.5$\times$ & 3.0$\times$ \\
TIES & 3.0$\times$ & 3.0$\times$ & 3.0$\times$ & 2.0$\times$ & 1.0$\times$ \\
DARE+TIES & 2.0$\times$ & 5.0$\times$ & 0.25$\times$ & 3.0$\times$ & 5.0$\times$ \\
Greedy Soup & 2.0$\times$ & 0.75$\times$ & 0.25$\times$ & 0.5$\times$ & 1.0$\times$ \\
\bottomrule
\end{tabular}%
}
\vspace{0.5ex}
\caption{Optimal training duration per domain for Qwen3.5-2B.}
\label{tab:optimal_2b}
\end{table}

\begin{table}[h]
\centering
\small
\setlength{\tabcolsep}{3pt}
\resizebox{\columnwidth}{!}{%
\begin{tabular}{lccccc}
\toprule
Method & Mathematics & Code & Instruction & Multilingual & Safety \\
\midrule
Simple Averaging & 0.25$\times$ & 0.75$\times$ & 0.5$\times$ & 0.25$\times$ & 1.0$\times$ \\
Task Arithmetic & 0.25$\times$ & 1.5$\times$ & 1.0$\times$ & 2.0$\times$ & 3.0$\times$ \\
TIES & 3.0$\times$ & 3.0$\times$ & 5.0$\times$ & 5.0$\times$ & 3.0$\times$ \\
DARE+TIES & 3.0$\times$ & 5.0$\times$ & 3.0$\times$ & 5.0$\times$ & 5.0$\times$ \\
Greedy Soup & 2.0$\times$ & 3.0$\times$ & 0.25$\times$ & 2.0$\times$ & 1.0$\times$ \\
\bottomrule
\end{tabular}%
}
\vspace{0.5ex}
\caption{Optimal training duration per domain for Qwen3.5-4B.}
\label{tab:optimal_4b}
\end{table}

\vspace*{10ex}

\end{document}